\documentclass{bmvc2k}

\title{Unsupervised Human Action Recognition with Skeletal Graph Laplacian and Self-Supervised Viewpoints Invariance}

\addauthor{Giancarlo Paoletti}{giancarlo.paoletti@iit.it}{1}
\addauthor{Jacopo Cavazza}{jacopo.cavazza@iit.it}{1}
\addauthor{Cigdem Beyan$^{1, }$}{cigdem.beyan@unitn.it}{2}
\addauthor{Alessio Del Bue}{alessio.delbue@iit.it}{1}

\addinstitution{
 Pattern Analysis and Computer Vision (PAVIS)\\
 Istituto Italiano di Tecnologia (IIT)\\
 Genova, Italy
}
\addinstitution{
 Department of Information Engineering and Computer Science\\
 University of Trento\\
 Trento, Italy
}

\runninghead{Paoletti \etal}{Unsupervised HAR with Graph Laplacian and SSVI}

\def\eg{\emph{e.g}\bmvaOneDot}

\def\etal{\emph{et al}\bmvaOneDot}
\def\ie{\emph{i.e}\bmvaOneDot}
\def\wrt{\emph{w.r.t}\bmvaOneDot}

\usepackage{amssymb}
\usepackage{enumitem}
\usepackage{multicol}
\usepackage{multirow}
\usepackage{overpic}
\usepackage{sidecap}
\usepackage{algpseudocode, algorithm, algorithmicx}
\usepackage{pifont}
\usepackage{wrapfig}
\usepackage[export]{adjustbox}

\algrenewcommand\algorithmicrequire{\textbf{Input:}}

\usepackage{xcolor}
\usepackage{tikz}
\usepackage{pgfplots}
\usetikzlibrary{shapes.geometric}
\pgfdeclareplotmark{mystar}{\node[star,star point ratio=2.25,minimum size=6pt,inner sep=0pt,draw=black,solid,fill=red] {};}
\pgfplotsset{width=\columnwidth, height=.5\columnwidth, compat=1.9}

\begin{document}
\maketitle

%-------------------------------------------------------------------------
\vspace{-5 mm}
\begin{abstract}
This paper presents a novel end-to-end method for the problem of skeleton-based unsupervised human action recognition.
We propose a new architecture with a convolutional autoencoder that uses graph Laplacian regularization to model the skeletal geometry across the temporal dynamics of actions.
Our approach is robust towards viewpoint variations by including a self-supervised gradient reverse layer that ensures generalization across camera views.
The proposed method is validated on NTU-60 and NTU-120 large-scale datasets in which it outperforms all prior unsupervised skeleton-based approaches on the cross-subject, cross-view, and cross-setup protocols.
Although unsupervised, our learnable representation allows our method even to surpass a few supervised skeleton-based action recognition methods.
The code is available in: \url{www.github.com/IIT-PAVIS/UHAR_Skeletal_Laplacian}
\end{abstract}

%-------------------------------------------------------------------------
\vspace{-6 mm}
\section{Introduction} \label{section:intro}

\begin{figure}[t] \centering
    \includegraphics[width=.55\columnwidth]{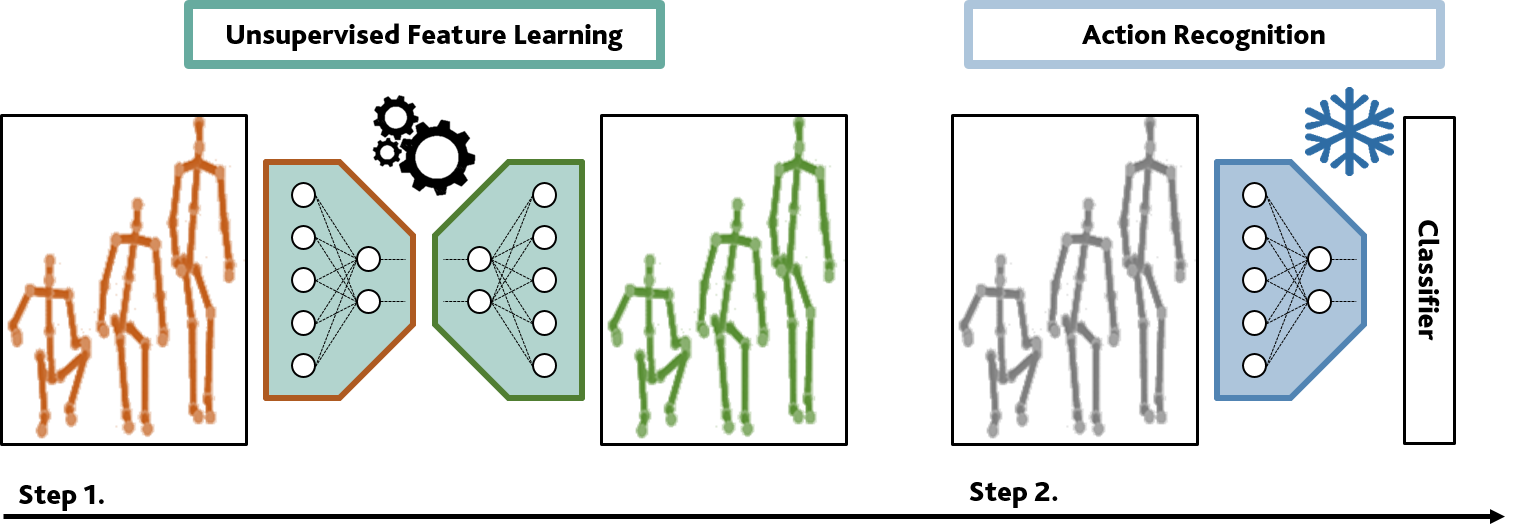}
    \vspace{-2 mm}
    \caption{{\bf Unsupervised Human Action Recognition (U-HAR) from skeleton data.} We compute features without using any supervision but by learning how to reconstruct skeleton data extracted with a generative approach. U-HAR evaluation relies on applying 1-Nearest Neighbor ($1$-NN) classifier or Linear Evaluation Protocol \cite{zheng2018unsupervised,Su2020,rao2020augmented,holden2015Siggraph,nie2020unsupervised,xu2020prototypical,kundu2018unsupervised,lin2020ms2l}.}
    \vspace{-5 mm}
    \label{figure:overall}
\end{figure}

\vspace{-3 mm}
Human Action Recognition (HAR) is ubiquitous across several computer vision applications, ranging from smart video surveillance, human-robot interaction, remote healthcare monitoring, and smart homes, to name a few.
The recent success in skeleton-based HAR, particularly by adopting deep learning methodologies, primarily relies on the supervised learning paradigm \cite{Cheng_2020_CVPR,Zhang_2017_ICCV,yan2018spatial}.
However, data annotation is expensive, time-consuming, and prone to human errors \cite{paoletti2020subspace}.
As a (recent) alternative, unsupervised approaches \cite{zheng2018unsupervised,Su2020,rao2020augmented,holden2015Siggraph,nie2020unsupervised,xu2020prototypical,kundu2018unsupervised,lin2020ms2l} are continuously reducing the performance gap with the fully supervised counterpart while dismissing the strong reliance over annotated data.

This paper tackles the unsupervised HAR (U-HAR) problem as formalized \eg in \cite{Su2020} and illustrated in Fig.~\ref{figure:overall}.
Using unannotated 3D skeleton sequences, we learn a feature representation, which is then fed to an action recognition classifier (\eg, 1-nearest neighbor) to validate the method performance as defined in standard evaluation protocols \cite{zheng2018unsupervised,Su2020,rao2020augmented,holden2015Siggraph,nie2020unsupervised,xu2020prototypical,kundu2018unsupervised,lin2020ms2l}.
We propose a novel unsupervised method for U-HAR that learns action representations through a \textbf{\textit{convolutional (residual) autoencoder}} (see Fig.~\ref{figure:pipeline}).
By doing so, we demonstrate the benefits of performing residual convolutions to jointly learn representations with spatio-temporal convolutions instead of relying on more complex and/or memory-intense architectures, which use \eg contrastive learning, GANs, gated networks, or recurrent networks \cite{zheng2018unsupervised,Su2020,rao2020augmented,holden2015Siggraph,nie2020unsupervised,xu2020prototypical,kundu2018unsupervised,lin2020ms2l}.

To boost the performance even further, we adopt \textbf{\textit{(graph) Laplacian regularization}} \cite{belkin2006manifoldreg} to learn representations that are aware of the spatial configuration of the \textit{skeletal geometry}.
We apply this regularization in the \textit{reconstruction} space (\ie, the space induced by the last layer of the decoder)
to inject a "continuity pattern" while making this "approximation" smoother.
This work is the first attempt where Laplacian Regularization is used within an unsupervised feature learning paradigm for action recognition.

To promote the deployment of our method in practical scenarios, we also tackle the problem of  viewpoint-invariance as camera positions and orientations used to capture humans very likely differ from the setup used in the tested dataset.
We improve \textbf{\textit{viewpoint-invariance}} by, first, perturbing the original data with random rotations.
Then, to increase the generalizability of the model, we enhance the unsupervised learned data representations by pairing the Laplacian-regularized reconstruction loss with a regressor head.
This regressor attempts to learn the parameters (rotation angles) of the random rotations we applied.
Using adversarial training in the form of a gradient reversal layer \cite{ganin2015unsupervised}, we learn a feature representation that can fool this regressor, being, thus, not influenced by the rotational perturbation.
This is a proxy for rotational invariance that we achieve with a different (and more effective - see Table \ref{table:GRL} and Section~\ref{section:transfer}) method than the Siamese network proposed in \cite{nie2020unsupervised} (only attempting to align rotated with non-rotated data).
It is important to notice that we do not achieve invariance towards some annotated features of the data, but we directly synthesize the random rotations, generated from the data itself: we thus leveraging on the concept of \textit{\textbf{self-supervision}}.

To validate our method, experiments were realized on two large-scale skeletal action datasets: NTU-60 (cross-subject and cross-view) \cite{Shahroudy_2016_NTURGBD} and NTU-120 (cross-subject and cross-setup) \cite{Liu_2019_NTURGBD120}.
Ablation studies are performed to dissect the impact of our autoencoder, the skeletal graph Laplacian, and our adaptation of gradient reversing to U-HAR.
The \emph{end-to-end} approach we proposed outperforms prior unsupervised skeleton-based methods for U-HAR.
It also favorably scores \wrt state-of-the-art supervised methods, even outperforming a few of them (see Fig.~\ref{figure:sota}).

%-------------------------------------------------------------------------
\vspace{-3 mm}
\section{Related Work} \label{section:relatedwork}

\vspace{-3 mm}
{\bf Unsupervised skeleton-based HAR.}
Encoder-decoder recurrent architectures are often used to solve HAR problems~\cite{kundu2018unsupervised,zheng2018unsupervised,lin2020ms2l,Su2020,rao2020augmented}.
Zheng \etal~\cite{zheng2018unsupervised} introduce LongT GAN, based on GRUs that learns how to represent skeletal body poses in time, with an adversarial loss supporting an auxiliary inpainting task.
\emph{MS\textsuperscript{2}L} \cite{lin2020ms2l} is also based on GRUs and benefits from contrastive learning, motion prediction, and jigsaw puzzle recognition.
In addition, Kundu \etal~\cite{kundu2018unsupervised} include a GAN-based encoder in their recurrent architecture (EnGAN).
{\em PCRP}~\cite{xu2020prototypical} builds upon a vanilla autoencoder trained to reconstruct the skeletal data using expectation maximization with learnable class prototypes.
Su \etal~\cite{Su2020} present the Predict \& Cluster (P\&C) method based on encoder-decoder RNN.
AS-CAL~\cite{rao2020augmented} combines contrastive learning with momentum LSTM where the similarity between augmented instances and the input skeleton sequence is contrasted, and then a momentum-based LSTM encodes the long-term actions.
SeBiReNet \cite{nie2020unsupervised} uses a Siamese denoising autoencoder is used with feature disentanglement, showing good performance across pose denoising and unsupervised cross-view HAR.
Unlike related works, our autoencoder is built on residual convolutions, showing the benefits of using simpler but superior architecture \wrt methods adapting gated or recurrent units, contrastive learning, and GANs.
Recently, Li \etal \cite{li2021crossclr} processed the joint, motion, and bone information altogether instead of using skeleton data supplied by the datasets.
Using these three modalities within a contrastive learning schema, this method improved the U-HAR results of NTU-60 cross-subject and cross-view (77.8\% and 83.4\% respectively), and NTU-120 cross-subject and cross-setup (67.9\% and 66.7\%, respectively).

%-------------------------------------------------------------------------
\noindent {\bf Laplacian Regularization.}
Belkin \etal \cite{belkin2006manifoldreg} propose to regularize a model using the implicit geometry of the feature space, regardless of the distribution of their labels, by using the Laplacian of the graph built over the cross-similarity of examples.
A similar approach was pursued by a recent end-to-end trainable approach for image denoising \cite{pang2017denoisingAE}.
We follow a different approach by applying Laplacian regularization in space while our autoencoder learns to reconstruct input skeletal data (\ie, \emph{reconstruction} space).
In this way, our goal is to inject the information of skeletal geometry into our model.
Different from \emph{supervised} HAR methods (\eg, \cite{liu2016spatiotemporal,yang2020centrality}) that directly exploit the "raw" adjacency matrix to encode skeletal connectivity, we take advantage of a more powerful mathematical tool, the graph Laplacian, since it better capitalizes from the skeletal geometry.
This differs from prior works, \eg \cite{zheng2018unsupervised,Su2020} relying on  Mean-Squared Error (MSE)-based action reconstruction only.

%-------------------------------------------------------------------------
\noindent {\bf Gradient Reversing.}
Originally proposed for domain adaptation, the gradient reversal layer (GRL) \cite{ganin2015unsupervised} is arguably useful to achieve a better generalization: \eg, classify actions performed by multiple subjects \cite{zunino2020predicting}.
Differently, we propose to endow a non-discriminative architecture (an autoencoder) for viewpoint-invariance.
We perform this by first synthesizing auxiliary rotations of the skeletal joints to simulate different viewpoints.
Then, by achieving the invariance across viewpoints by a GRL layer that is fooling a predictor attempting to infer the viewpoint from the hidden representation of our autoencoder.
Li \etal \cite{li2018unsupervised} adapts GRL to obtain view-invariant action representations.
However, that work differs from ours by (a) relying on RGB-D data and, more importantly, (b) using the annotated viewpoints of the datasets as source and target domains and learn how to distinguish them by classification.

%-------------------------------------------------------------------------
\vspace{-4 mm}
\section{Proposed Method} \label{section:method}
\begin{figure}[t] \centering
    \includegraphics[width=.6\columnwidth, valign=c]{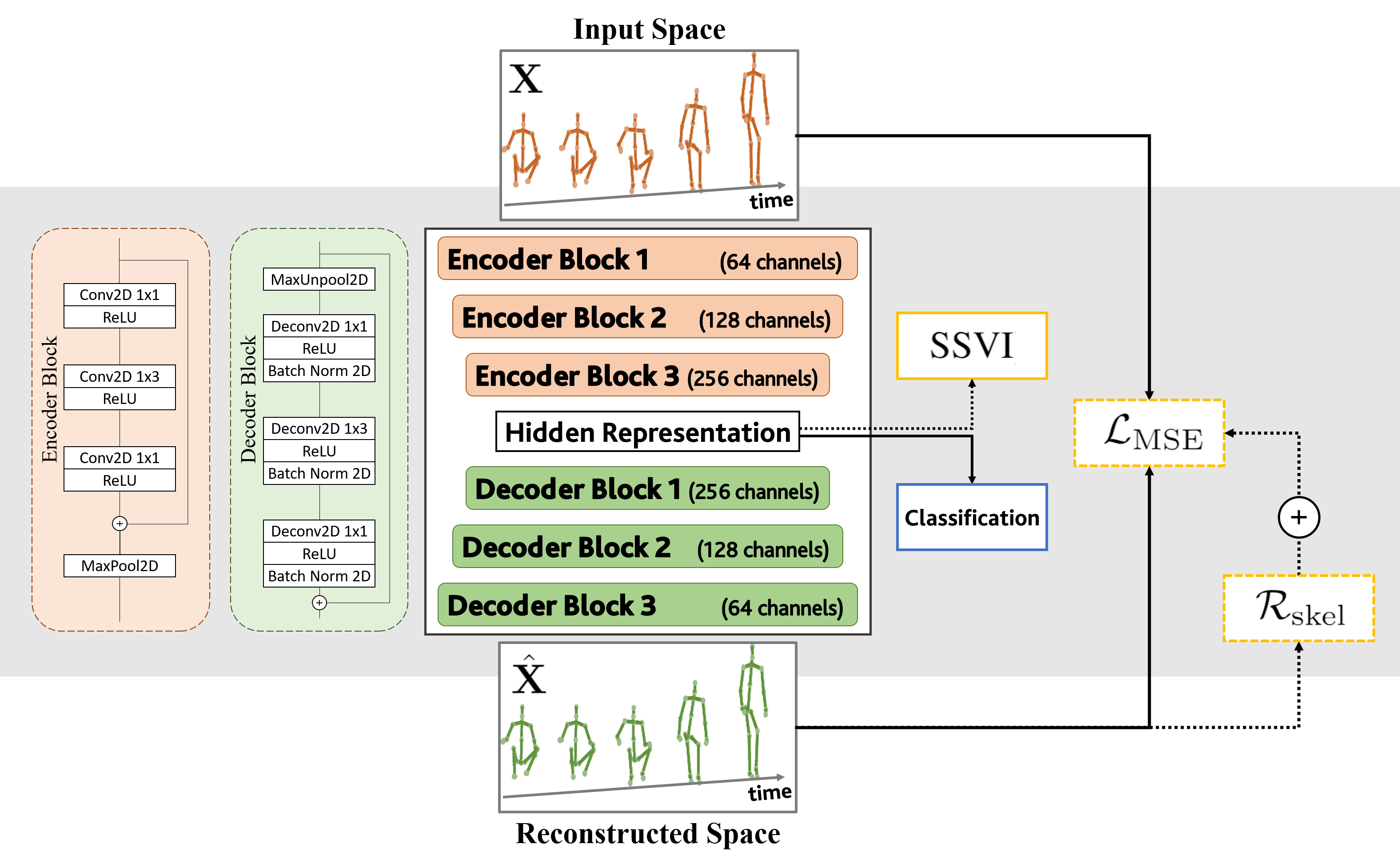}
    \hfill
    $\vcenter{\hbox{\begin{subfigure}{} \scalebox{.55}{
        \begin{minipage}{.65\linewidth}
            \begin{algorithm}[H]
                \caption{Training our proposed approach}
                \begin{algorithmic}[1]
                    \State Randomly initialize $\mathbf{E}_\varphi$, $\mathbf{D}_\theta$ and the SSVI module.
                    \State Compute the skeletal graph Laplacian $\mathbf{L}$.
                    \While{not converged}
                        \State Sample a mini-batch of data $\mathcal{B}$.
                        \State Do a forward pass through $\mathbf{E}_\varphi$ and $\mathbf{D}_\theta$.
                        \State Update $\mathbf{E}_\varphi$, $\mathbf{D}_\theta$ using the MSE loss as in Eq~\eqref{equation:MSEloss}. \quad \quad
                        \Comment{\scalebox{0.8}{\textsc{(Optional Skeletal Laplacian Regularization)}}}
                        \State Update $\mathbf{E}_\varphi$, $\mathbf{D}_\theta$ using the $\mathcal{R}_{\rm skel}$ loss as in Eq~\eqref{equation:skeleton_laplacian_regularizer}. \quad \quad
                        \Comment{\scalebox{0.8}{\textsc{(Optional Viewpoints Invariance)}}}
                        \State Randomly sample $\alpha,\beta,\gamma$ in $[0,2\pi]$
                        \State Rotate all data in $\mathcal{B} \longrightarrow \mathcal{B}^{(\alpha,\beta,\gamma)}$
                        \State Do a forward pass through $\mathbf{E}_\varphi$
                        \State Update $\mathbf{E}_\varphi$ using the SSVI module fed by $\mathcal{B}^{(\alpha,\beta,\gamma)}$.
                    \EndWhile
                \end{algorithmic}
                \label{algorithm:pseudo-code}
            \end{algorithm}
        \end{minipage}}
    \end{subfigure}}}$
    \vspace{2 mm}
    \caption{{\bf Our proposed method (Left):} we exploit a convolutional autoencoder (\textbf{AE}) trained with $\mathcal{L}_{\rm MSE}$ (Eq.~\eqref{equation:MSEloss}). In the reconstruction space, we perform a \textit{Skeletal Laplacian Regularization} (\textbf{L}; Sec.~\ref{section:laplacian}, Eq.~\eqref{equation:skeleton_laplacian_regularizer}), enriching the learned (hidden) feature representations with the skeletal geometry information. We also include a \textit{self-supervised viewpoint-invariance} (\textbf{SSVI} module, Sec.~\ref{section:GRL}), which adapts a gradient reversal layer \cite{ganin2015unsupervised} to achieve robustness towards different viewpoints. Our convolutional encoder and deconvolutional decoder blocks both exploit residual connections while batch normalization is exclusive for the decoder. {\bf Right:} Pseudo-code of the training process.}
    \vspace{-5mm}
    \label{figure:pipeline}
\end{figure}
\vspace{-3 mm}
We present our unsupervised approach by introducing the proposed convolutional autoencoder (Section \ref{section:convae}) and the Laplacian regularization (Section \ref{section:laplacian}).
Following that, we discuss the self-supervised viewpoint invariance module (SSVI; Section \ref{section:GRL}).
%-------------------------------------------------------------------------
\vspace{-3 mm}
\subsection{Convolutional Autoencoder} \label{section:convae}
\vspace{-2 mm}
The proposed Convolutional Autoencoder (\textbf{AE}) input is a set of 3D human body joints in time extracted from a video sequence with one or more subjects performing an unlabelled action.
Let $\mathbf{X}$ denote an input sequence of body joints represented as a $d \times m \times t$ tensor, containing the $x,y,z$ coordinates ($d =3$), the number of joints ($m=25$ on NTU-60 \cite{Shahroudy_2016_NTURGBD} and NTU-120 \cite{Liu_2019_NTURGBD120}) and the number of timestamps $t$\footnote{To be comparable with prior art, we cast each skeleton sequence to a fixed temporal length \cite{Su2020}.}.
We aim at obtaining \textit{unsupervised feature representations} by learning an autoencoder that reconstructs the input data $\mathbf{X}$ using a Mean-Squared Error (MSE) loss:
\vspace{-3mm}
\begin{equation}\label{equation:MSEloss}
    \mathcal{L}_{\rm MSE} = \tfrac{1}{2} \mathbb{E}_{\mathbf{X} \sim \mathcal{B}} \big[\| \mathbf{X} - \hat{\mathbf{X}} \|_F^2 \big],
    \vspace{-2mm}
\end{equation}
where $\| \cdot \|_F$ denotes the Frobenius norm, \ie, the Euclidean norm of the vector obtained after flattening the tensor.
The MSE loss in Eq.~\eqref{equation:MSEloss} is minimized by using gradient descent (Adam optimizer) over mini-batches $\mathcal{B}$.
The reconstructed data are defined as \ $\hat{\mathbf{X}} = \mathbf{D}_\theta \circ \mathbf{E}_\varphi (\mathbf{X})$ \ 
and computed using an encoder-decoder architecture, where $\varphi$ denotes the learnable parameters of the encoder $\mathbf{E}$ and $\theta$ are the analogous parameters for the decoder $\mathbf{D}$.
The complete architecture of our convolutional autoencoder is detailed in Fig.~\ref{figure:pipeline}.
We concatenate different residual blocks to build the autoencoder: 3 for the encoder and 3 for the decoder. 
At the end of the encoder, a FC layer represents the latent space $\mathbf{z}$ of size 2048.
The size of $\mathbf{z}$ was determined by testing various numerical combinations, \eg, 32, 128, 512.
For our convolutional autoencoder, 2048 results in the best performances (up to +10\% in NTU-60 and +23\% in NTU-120) out of all combinations. Thus, this value was fixed in all experiments. \\
\noindent
\textit{\textbf{Residual blocks of convolutions.}}  Our \textbf{AE} architecture stacks different fully-residual blocks for both encoder and decoder, whereas each block is made of convolutions capable to jointly learning spatial representations of skeletal data in time, treating each skeletal data $\mathbf{X}$ as 2D convolutions.
Convolutions with fixed size kernels (either $1\times1$ or $1\times3$), applied inside $\mathbf{E}$ and $\mathbf{D}$, are capable to capture spatial and temporal relationships of data along tensor rows for the former and along tensor columns for the latter. Hence it is called \emph{convolutions-in-time}.
In detail, within the encoder blocks, the residual layer is made of a series of three \emph{2D-convolutional} layers (each with \emph{ReLU} activations) stacked together.
At the same time, decoder blocks share a similar structure but using instead \emph{2D-deconvolutional} layers with the addition of \emph{2D-BatchNorm} applied after each \emph{ReLU} activation.
To ensure the bottleneck structure of the convolutional autoencoder, a \emph{MaxPool} layer is applied at the end of each encoder block, whereas a \emph{MaxUnpool} layer is applied at the beginning of each decoder block (see Fig.~\ref{figure:pipeline}).

%-------------------------------------------------------------------------
\vspace{-3 mm}
\subsection{Skeletal Laplacian Regularization} \label{section:laplacian}
\vspace{-2 mm}
The graph Laplacian is an established tool to analyze weighted undirected graphs.
It builds upon the graph adjacency matrix $\mathbf{W}$, whose entries $W_{ij}$ are defined such that $W_{ij} = 1$ if and only if the nodes $i$ and $j$ are connected through an edge.
The (un-normalized) graph Laplacian $\mathbf{L}$ is easily computable from $\mathbf{W}$ as $\mathbf{L} = \mathbf{D} - \mathbf{W}$, where $\mathbf{D}$ is the degree matrix (obtained as the diagonal matrix where its $(i,i)$-th element is $D_{ii} = \sum_j W_{ij}$) \cite{deo1974graph}.
The Laplacian regularizer $\mathcal{R}(\mathbf{z}) = \sum_{i,j} W_{ij} (z_i - z_j)^2$
can be applied to a hidden vectorial embedding $\mathbf{z}$ to learn the geometry of the feature space (where $\mathbf{z}$ belongs to) and to capitalize from these cues to solve a semi-supervised learning paradigm \cite{belkin2006manifoldreg}.
This is true because, thanks to the weights $W_{ij}$, we can prioritize the alignment between the scalar components $z_{i}$ and $z_{j}$ by simply putting a stronger penalty between pairs of components that must be well aligned.
In our case, we attempt to do so by promoting the alignment of skeletal joints, which are connected through a bone (\eg, \textit{an edge exists if and only if joints are connected}).
The results given in Supplementary Material show that such a setting is also empirically favorable compared to other ways of initializing $\mathbf{W}$.
We intend this as a valid proxy for injecting the knowledge of skeletal geometry while learning our action representations.
The reason why $\mathcal{R}$
is termed Laplacian regularizer lies in the fact that $\mathcal{R}(\mathbf{z}) = 2 \ \mathbf{z}^\top \mathbf{L} \mathbf{z}$.
That is, $\mathcal{R}(\mathbf{z})$ implements a "$\mathbf{L}$-weighted weight decay" - since $\mathcal{R}(\mathbf{z}) = \|\mathbf{Q}\mathbf{z}\|_2^2$ if we set $\mathbf{Q} = \sqrt{\mathbf{L}}$.

Unlike prior art \cite{belkin2006manifoldreg,pang2017denoisingAE}, we apply Laplacian regularization to the \textbf{\textit{reconstruction space}} learned by our decoder, \ie, the space where $\hat{\mathbf{X}}$ belongs to.
We compute our proposed \textbf{\textit{skeletal Laplacian regularizer}} as:
\vspace{-4 mm}
\begin{equation}\label{equation:skeleton_laplacian_regularizer}
    \mathcal{R}_{\rm skel} = \mathbb{E}_{\rm \mathbf{X} \sim \mathcal{B}} \Big[ \mathbb{E}_{t,d} \big[ {{\hat{\mathbf{x}}^{(t,d)}}}{^\top} \mathbf{L} \hat{\mathbf{x}}^{(t,d)}\big] \Big],
    \vspace{-4 mm}
\end{equation}
where $\hat{\mathbf{x}}^{(t,d)}$ is the $m$-dimensional column vector stacking the scalar (absciss\ae, ordinat\ae~or quot\ae) coordinates along the dimension $d$ obtained from the reconstructed sequence $\hat{\mathbf{X}}$ at time $t$.
In Eq.~\eqref{equation:skeleton_laplacian_regularizer}, the regularizer $\mathcal{R}_{\rm skel}$ is averaged over the mini-batch $\mathcal{B}$, considering the reconstructions produced by the convolutional autoencoder across coordinates and timestamps.
The Laplacian regularization attempts to inject the connectivity of the skeleton to learn a feature representation, which is aware of the \textbf{\textit{skeletal geometry}}.
We deem this to be a proxy of features that are aware of the fact that the representation learned, \eg, from the shoulder and elbow joints, cannot be decorrelated from each other since those joints are closed in space, while there can be joints, which are more distant in space (\eg, left foot vs. right hand) are allowed to be more independent.

%-------------------------------------------------------------------------
\vspace{-3 mm}
\subsection{Self-supervised Viewpoints Invariance (SSVI)} \label{section:GRL}

\begin{figure}[t] \centering
    \includegraphics[width=.4\columnwidth, valign=c]{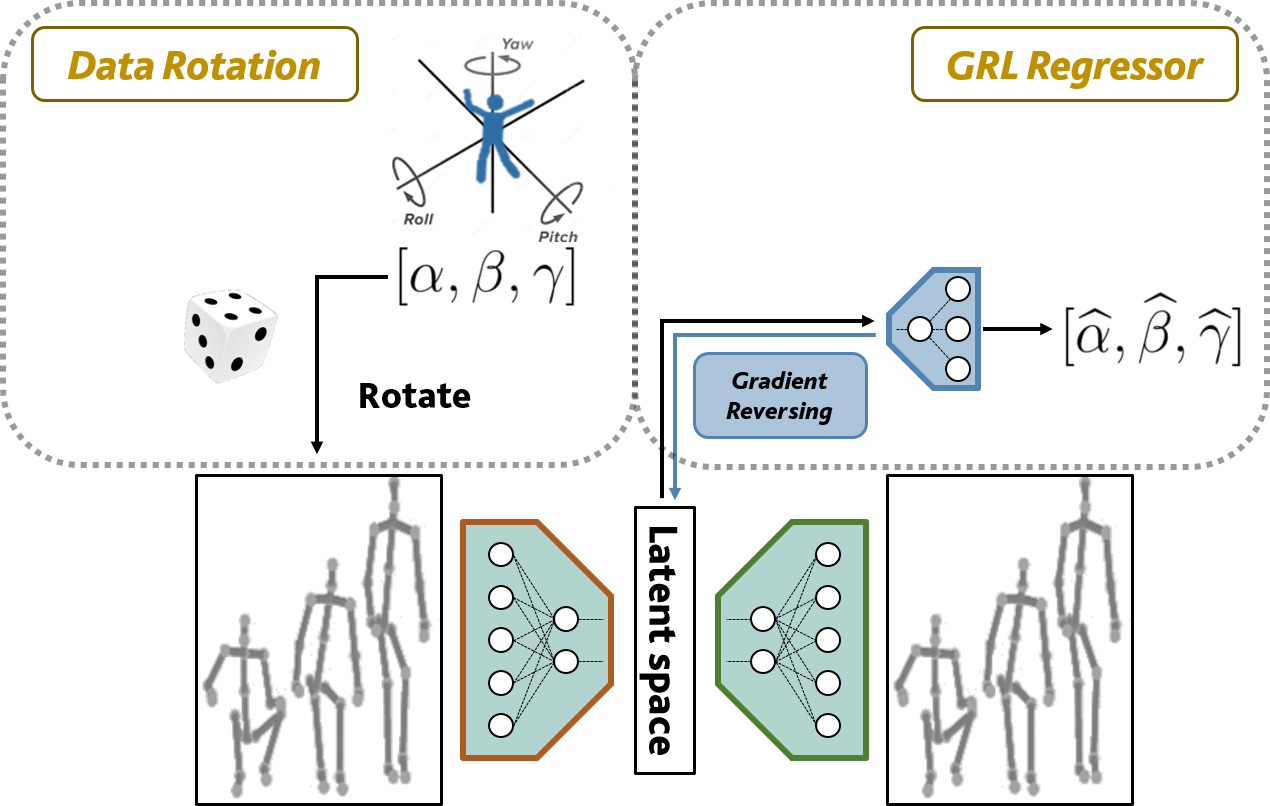}
    \vspace{3 mm}
    \caption{{\bf Self-Supervised Viewpoints Invariance} using a regressor and a gradient reversal layer \cite{ganin2015unsupervised}. We enforce the hidden representation, learned by our encoder to be invariant across synthetic rotations that we applied to the input data $\mathbf{X}$, using the Euler's angles $\alpha,\beta,\gamma$. We claim it to be the proxy to achieve viewpoints invariance and generalize across random rotations (parametrized by Euler's angles $\alpha,\beta,\gamma$).}
    \vspace{-4 mm}
    \label{figure:SSVI}
\end{figure}

\vspace{-2 mm}
We propose to obtain a viewpoint-invariant action representation by first synthesizing multiple viewpoints of the original skeletal data.
Geometrically, this operation can be easily framed as (right) multiplying $\mathbf{X}^t$, the $m \times 3$ matrix stacking the $m$ 3D joints captured at a given timestamp $t$, by $\boldsymbol{\Omega}$ defined as the product of $\boldsymbol{\Omega}_x$, $\boldsymbol{\Omega}_y$, and $\boldsymbol{\Omega}_z$, each corresponding to the independent three (planar) rotations performed around the $x,y,z$ axis, respectively.
$\boldsymbol{\Omega}_x$ depends upon the pitch angle $\alpha$, $\boldsymbol{\Omega}_y$ depends upon the yaw angle $\beta$, and $\boldsymbol{\Omega}_z$ depends upon the roll angle $\gamma$.
By means of the so-defined $\boldsymbol{\Omega}$, we can obtain $\mathbf{Z}^t = \mathbf{X}^t\boldsymbol{\Omega},$ and, hence, \textbf{\textit{synthesize}} a rotation under a \textbf{\textit{generated viewpoint}} by iterating the process over all timestamps $t$ of the sequence $\mathbf{X}$ and, afterward repeating the whole procedure for all sequences $\mathbf{X}$ in the mini-batch $\mathcal{B}$, we generate transformed sequences $\mathbf{Z}$.
When $\mathbf{Z}$ is obtained from $\mathbf{X}$ according to this procedure, the action class referring to them remains unaltered in its information content, while only the viewpoint had changed.

We make $\mathbf{Z}$ and $\mathbf{X}$ indistinguishable, being the latter a proxy for an improved hidden representation that our autoencoder learns from data, since in this way, the autoencoder will be robust towards different viewpoints, and we purport that this requirement is a proxy for an improved viewpoint generalization.
We use a L1 norm to train a \textbf{\emph{regressor}} that predicts the triplet $[\alpha,\beta,\gamma]$ that is used to rotate the data.
We take advantage of a \textbf{\textit{gradient reversal layer (GRL)}} \cite{ganin2015unsupervised} to flip the gradients coming from the regressor.
By doing so, we make sure to actively promote invariance across synthetic rotations by explicitly optimize the learned representation to fool a regressor that is attempting to predict the $[\alpha,\beta,\gamma]$ triplet used to rotate the data of each mini-batch before every forward pass.
We call this the \textbf{\textit{self-supervised viewpoints invariance} (SSVI)} module, which is visualized in Fig.~\ref{figure:SSVI} and connected to the hidden representation of the autoencoder (see Fig.~\ref{figure:pipeline}).

%-------------------------------------------------------------------------
\vspace{-3 mm}
\subsection{Pseudo-code \& Inference} \label{section:inference}

\vspace{-2 mm}
The pseudo-code of our method is given in Algorithm \ref{algorithm:pseudo-code} (see Fig.~\ref{figure:pipeline} Right).
Below, we provide details on how a trained autoencoder is used for inference.
All implementation details, including learning curves, can be found in the Supplementary Material. \\

%---------------------------------------------------------------
\textbf{Inference.}
Standard evaluation protocols for U-HAR perform inference by keeping the learned representations frozen and training a classifier on top of them.
Two alternatives are used: a \emph{linear classifier} using the available labels of the datasets (\emph{\textbf{Linear Evaluation Protocol (LEP)}}) \cite{zheng2018unsupervised,rao2020augmented,kundu2018unsupervised, holden2015Siggraph,nie2020unsupervised,xu2020prototypical}
and a $1$-nearest neighbor predictor (\emph{\textbf{$1$-NN}}) \cite{Su2020}.

%-------------------------------------------------------------------------
\vspace{-3 mm}
\section{Experimental Analysis} \label{section:results}
\vspace{-3 mm}
The proposed method is validated using the two large-scale skeletal action datasets: NTU-60 \cite{Shahroudy_2016_NTURGBD} and NTU-120 \cite{Liu_2019_NTURGBD120}.
\textbf{NTU-60} dataset contains 60 action classes performed by 40 subjects, captured with Microsoft Kinect v2 (25 joints).
\textbf{NTU-120} encompasses 120 action classes of 106 subjects while there are in total 32 different setups (\eg, different backgrounds or locations where the data is captured).
We evaluate the proposed method on NTU-60 dataset for cross-subject (C-Subject) and cross-view (C-View) settings \cite{Shahroudy_2016_NTURGBD}, and NTU-120 for cross-subject (C-Subject) and cross-setup (C-Setup) settings \cite{Liu_2019_NTURGBD120}.

%-------------------------------------------------------------------------
\vspace{-3 mm}
\subsection{Comparisons against the state-of-the-art}
\vspace{-2 mm}
Herein, we discuss the improvements that our proposed approach (AE-L) brings in, that is summarized as: for $1$-NN Protocol, AE-L performs +3.4\% and +6.8\% on NTU-60 C-Subject and C-View, respectively, and +0.7\% and +2.0\% on NTU-120 C-Subject and C-Setup, respectively, over \cite{Su2020}.
For LEP, we improve prior art on NTU-60 C-Subject (+1.3\%), NTU-60 C-View (+7.6\%), on NTU-120 C-Subject (+10.5\%) and on NTU-120 C-Setup (+13.2\%).
Detailed discussion is provided below.
\begin{table}[t] \centering
    \resizebox{.4\columnwidth}{!}{
        \begin{tabular}[t]{lcc} \hline
            \multirow{2}{*}{\bf NTU-60 \cite{Shahroudy_2016_NTURGBD}} & C-Subject                          & C-View                             \\
                                                                      & ACC (\%)                           & ACC (\%)                           \\ \hline
            \multicolumn{3}{c}{\bf{$1$-NN Protocol} \cite{Su2020}} \\ \hline
            P\&C FS$^\star$ \cite{Su2020}                             & 50.6                               & 76.3                               \\
            P\&C FW$^\star$ \cite{Su2020}                             & 50.7                               & 76.1                               \\
            \textit{Baseline} {\bf AE}                                & 50.1                               & 80.4                               \\
            \textit{Our} {\bf AE}                                     & \underline{\textit{52.3}}          & \underline{\textit{81.0}}          \\
            \textit{Our} {\bf AE-L} (AE + $\mathcal{R}_{\rm skel}$)   & \underline{\textit{\textbf{54.1}}} & \underline{\textit{\textbf{83.1}}} \\ \hline
            \multicolumn{3}{c}{\bf{Linear Evaluation Protocol (LEP)} \cite{zheng2018unsupervised}} \\ \hline
            LongT GAN \cite{zheng2018unsupervised}                    & 39.1                               & 48.1                               \\
            MS\textsuperscript{2}L \cite{lin2020ms2l}                 & 52.5                               & --                                 \\
            PCRP \cite{xu2020prototypical}                            & 53.9                               & 63.5                               \\
            VAE-PoseRNN \cite{kundu2018unsupervised}                  & 56.4                               & 63.8                               \\
            AS-CAL \cite{rao2020augmented}                            & 58.5                               & 64.6                               \\
            MM-AE \cite{holden2015Siggraph}                           & 61.2                               & 70.2                               \\
            EnGAN-PoseRNN \cite{kundu2018unsupervised}                & 68.6                               & 77.8                               \\
            SkeletonCLR joint \cite{li2021crossclr}                   & 68.3                               & 76.4                               \\
            \textit{Baseline} {\bf AE}                                & 68.5                               & 84.3                               \\
            \textit{Our} {\bf AE}                                     & \underline{\textit{69.2}}          & \underline{\textit{85.1}}          \\
            \textit{Our} {\bf AE-L} (AE + $\mathcal{R}_{\rm skel}$)   & \underline{\textit{\textbf{69.9}}} & \underline{\textit{\textbf{85.4}}} \\ \hline
        \end{tabular}}
    \quad
    \resizebox{.40\columnwidth}{!}{
        \begin{tabular}[t]{lcc} \hline 
            \multirow{2}{*}{\bf NTU-120 \cite{Liu_2019_NTURGBD120}} & C-Subject                          & C-Setup                            \\
                                                                    & ACC (\%)                           & ACC (\%)                           \\ \hline
            \multicolumn{3}{c}{\bf{$1$-NN Protocol} \cite{Su2020}} \\ \hline
            P\&C$^\dagger$ \cite{Su2020}                            & 41.7                               & 42.7                               \\
            \textit{Baseline} {\bf AE}                              & 40.2                               & 44.3                               \\
            \textit{Our} {\bf AE}                                   & \textit{41.0}                      & \underline{\textit{44.5}}          \\
            \textit{Our} {\bf AE-L} (AE + $\mathcal{R}_{\rm skel}$) & \underline{\textit{\textbf{42.4}}} & \underline{\textit{\textbf{44.7}}} \\ \hline
            \multicolumn{3}{c}{\bf{Linear Evaluation Protocol (LEP)} \cite{zheng2018unsupervised}} \\ \hline
            PCRP \cite{xu2020prototypical}                          & 41.7                               & 45.1                               \\
            AS-CAL \cite{rao2020augmented}                          & 48.6                               & 49.2                               \\
            \textit{Baseline} {\bf AE}                              & 56.4                               & 60.3                               \\
            \textit{Our} {\bf AE}                                   & \underline{\textit{57.1}}          & \underline{\textit{61.8}}          \\
            \textit{Our} {\bf AE-L} (AE + $\mathcal{R}_{\rm skel}$) & \underline{\textit{\textbf{59.1}}} & \underline{\textit{\textbf{62.4}}} \\ \hline
        \end{tabular}}
    \vspace{2 mm}
    \caption{Evaluation on NTU-60 \cite{Shahroudy_2016_NTURGBD} and NTU-120 \cite{Liu_2019_NTURGBD120}.
    Our numbers are in italic, an improved performance over prior art is underlined. The best of all results are in black. "Baseline AE" stands for the proposed AE without residual layers. $^\star$FS and FW stand for a decoder with "fixed states" and "fixed weights", respectively \cite{Su2020}. $^\dagger$Taken from PCRP \cite{xu2020prototypical}.}
    \vspace{-6.5mm}
    \label{table:NTUs_sota}
\end{table}
%-------------------------------------------------------------------------
\\
\textbf{Cross-subject evaluation protocol.}
We compare our AE-L against state-of-the-art methods (SOTA) using
recommended training and testing splits of NTU-60 \cite{Shahroudy_2016_NTURGBD} and NTU-120 \cite{Liu_2019_NTURGBD120} datasets.
We use only skeletal data in our experiments, \ie, we discard the RGB and depth images, normalizing data as in prior works \cite{Su2020}, and we feed our $\mathcal{R}_{\rm skel}$-regularized autoencoder (AE-L) with \textit{training data only} (\ie, with the data of subjects in training).
Then, we apply one of the two evaluations: $1$-NN or LEP, as described in Section \ref{section:inference}.
Readers can refer to Table \ref{table:NTUs_sota} (C-Subject columns) for the results of the analysis mentioned above.

Ablation study shows that AE-L improves the performance of AE model, demonstrating the advantages of using Laplacian regularization: +1.8\% in $1$-NN, +0.7\% in LEP for NTU-60 C-Subject and +1.4\% in $1$-NN, +2\% in LEP for NTU-120 C-Subject setting.
In addition, the usage of Laplacian regularization grants at least a +5\% performance gain over different action classes for both NTU-60 C-Subject and C-View, and NTU-120 C-Subject and C-Setup settings (the complete list of these action classes can be found in Supplementary Material).
Our AE is preferable compared to the baseline AE (i.e., not using residual layers in our design) as performing +2.2\% in $1$-NN, +0.7\% in LEP for NTU-60 C-Subject, and +0.8\% in $1$-NN, +0.7\% in LEP for NTU-120 C-Subject, showing the contribution of using residual convolutions layers.

For NTU-60 C-Subject, the learned features of our AE-L and AE models are superior to P\&C \cite{Su2020}:
+3.5\% as compared to P\&C FS \cite{Su2020} and +3.4\% as compared to P\&C FW \cite{Su2020}.
While exploiting LEP, our AE-L again performs better than the approaches based on RNNs \cite{kundu2018unsupervised,rao2020augmented}, performing +11.4\% better than AS-CAL \cite{rao2020augmented} and +8.7\% than MM-AE \cite{holden2015Siggraph}.
We improve VAE-PoseRNN \cite{kundu2018unsupervised}, EnGAN-PoseRNN \cite{kundu2018unsupervised} and SkeletonCLR joint \cite{li2021crossclr} by +13.5\%, +1.3\%, +1.6\%, respectively.
We also surpass MS\textsuperscript{2}L \cite{lin2020ms2l} (+17.4\%).
For NTU-120 C-Subject, AE-L outperforms P\&C \cite{Su2020} (+0.7\%) when $1$-NN is applied, with an increase in performance \wrt both AS-CAL \cite{rao2020augmented} (+10.5\%) and PCRP \cite{xu2020prototypical} (+17.4\%) in LEP.
%-------------------------------------------------------------------------
\\
\textbf{Cross-view and cross-setup evaluation protocols.}
We compare our AE-L against prior methods on NTU-60 \cite{Shahroudy_2016_NTURGBD} C-View and NTU-120 \cite{Liu_2019_NTURGBD120} C-Setup settings (see Table \ref{table:NTUs_sota}, C-View and C-Setup columns).
For NTU-60 \cite{Shahroudy_2016_NTURGBD} C-View, AE-L improves the performance by +6.8\% and +7.0\% over P\&C FS \cite{Su2020} and P\&C FW \cite{Su2020}, respectively within the $1$-NN Protocol.
In the same protocol, ablation study shows that AE-L improves the performance of Baseline AE by +2.1\%, and using residual layers (i.e., our AE) performs 0.6\% better than not using (Baseline AE).
On NTU-60 \cite{Shahroudy_2016_NTURGBD} C-View, with LEP, the superiority of AE-L is much visible such that it notably exceeds LongT GAN \cite{zheng2018unsupervised} (+37.3\%), PCRP \cite{xu2020prototypical} (+21.9\%), AS-CAL (+20.8\%), VAE-PoseRNN (+21.6\%), MM-AE (+15.2\%), EnGAN-PoseRNN (+7.6\%) and SkeletonCLR joint \cite{li2021crossclr} (+9\%).
Our AE also surpasses "Baseline AE" by 0.8\%, once again showing the positive contribution of residual layers.
On NTU-120 \cite{Liu_2019_NTURGBD120} C-Setup, AE-L again performs better than P\&C within the $1$-NN Protocol (+2.0\%), and in LEP, it performs better than AS-CAL and PCRP by margins of +13.2\% and +17.3\%, respectively.
In this setting, our AE achieves better results than "Baseline AE" by +0.2\% for 1-NN and +1.5\% for LEP.
\textbf{Confusion matrices} belonging to AE-L in testing can be found in Supplementary Material.

%-------------------------------------------------------------------------
\vspace{-3 mm}
\subsubsection{Comparisons Against Supervised Methods}
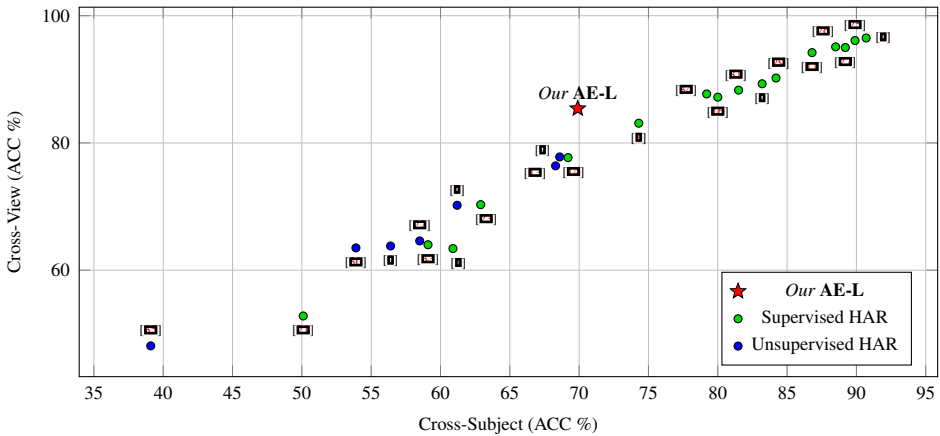
\begin{figure}[t] \centering
\begin{tikzpicture}
    \begin{axis}[scatter/classes={r={mark=mystar, fill=red}, draw=black,
                                  g={mark=*, fill=green!85!black, draw=black,solid, mark size=1.5pt},
                                  b={mark=*, fill=blue, draw=black,solid, mark size=1.5pt}},
                 xlabel=\scriptsize{{Cross-Subject (ACC \%)}},
                 ylabel=\scriptsize{{Cross-View (ACC \%)}},
                 yticklabel style= {font=\scriptsize},
                 xticklabel style= {font=\scriptsize},
                 legend pos=south east,
                 grid=both,
                 grid style={line width=.1pt, draw=gray!50},
                 nodes near coords*={\Label},
                 visualization depends on={value \thisrow{reference} \as \Label},
                 visualization depends on={value \thisrow{anchor}    \as \myanchor},
                 every node near coord/.append style={anchor=\myanchor}]
    \addplot[scatter, only marks, scatter src=explicit symbolic,]
        table[meta=class] {
            x    y    class reference                                       anchor
            50.1 52.8 g     \tiny{\cite{Rahmani2016}}               90
            59.1 64.0 g     \tiny{\cite{Du2015}}                    90
            60.9 63.4 g     \tiny{\cite{cavazza2019}}               110
            62.9 70.3 g     \tiny{\cite{Shahroudy2016}}             110
            69.2 77.7 g     \tiny{\cite{liu2016spatiotemporal}}     110
            74.3 83.1 g     \tiny{\cite{Kim2017}}                   90
            79.2 87.7 g     \tiny{\cite{Zhang_2017_ICCV}}           350
            80.0 87.2 g     \tiny{\cite{Mengyuan2017}}              90
            81.5 88.3 g     \tiny{\cite{yan2018spatial}}            280
            83.2 89.3 g     \tiny{\cite{ChaoLi2017}}                90
            84.2 90.2 g     \tiny{\cite{Wen_Gao_Fu_Zhang_Xia_2019}} 260
            86.8 94.2 g     \tiny{\cite{LiChenChenZhang2019}}       90
            88.5 95.1 g     \tiny{\cite{shi2019}}                   310
            89.2 95.0 g     \tiny{\cite{Si_2019_CVPR}}              90
            89.9 96.1 g     \tiny{\cite{Shi_2019_CVPR}}             270
            90.7 96.5 g     \tiny{\cite{cheng2020skeleton}}         180
            69.9 85.4 r     \scriptsize{\it{Our}\ {\bf AE-L}}           270
            39.1 48.1 b     \tiny{\cite{zheng2018unsupervised}}     270
            53.9 63.5 b     \tiny{\cite{xu2020prototypical}}        90
            56.4 63.8 b     \tiny{\cite{kundu2018unsupervised}}     90
            58.5 64.6 b     \tiny{\cite{rao2020augmented}}          270
            61.2 70.2 b     \tiny{\cite{holden2015Siggraph}}        270
            68.6 77.8 b     \tiny{\cite{kundu2018unsupervised}}     340
            68.3 76.4 b     \tiny{\cite{li2021crossclr}}            20
            };
    \legend{\scriptsize{\it{Our} {\bf AE-L}}, ,
            \scriptsize{Supervised HAR},
            \scriptsize{Unsupervised HAR}}
    \end{axis}
\end{tikzpicture}
\vspace{-4mm}
\caption{Comparisons between our AE-L and SOTA unsupervised and supervised skeleton-based HAR methods on NTU-60 dataset \cite{Shahroudy_2016_NTURGBD}.}
\vspace{-5 mm}
\label{figure:sota}
\end{figure}
\vspace{-2 mm}
We compare the performance of our AE-L with SOTA supervised skeleton-based HAR approaches on NTU-60 dataset \cite{Shahroudy_2016_NTURGBD}.
This comparison includes kernel-based methods \cite{Rahmani2016,cavazza2019} and the methods realizing feature learning \cite{Du2015,liu2016spatiotemporal,Shahroudy2016,Zhang_2017_ICCV,Kim2017,Mengyuan2017,ChaoLi2017,yan2018spatial,Wen_Gao_Fu_Zhang_Xia_2019,LiChenChenZhang2019,shi2019,Si_2019_CVPR,Shi_2019_CVPR,cheng2020skeleton} with several different deep learning architectures, \eg, RNNs, LSTMs, CNNs, and Graph Convolutional Networks (GCNs).
The corresponding results are presented in Fig.~\ref{figure:sota}, while an in-depth comparison is given in the Supplementary Material.

Our AE-L, although based on unsupervised learning, is able to achieve better performance than the fully supervised kernel-based methods \cite{Rahmani2016,cavazza2019}, with a +7.2\% to +19.8\% improvement in C-Subject and a +22\% to +32.6\% improvement in C-View setting.
AE-L also outperforms several fully supervised deep architectural methods: hierarchical RNN \cite{Du2015} (providing an increase of 10.8\% in C-Subject and up to 21.4\% in C-View), spatial-temporal LSTM \cite{liu2016spatiotemporal} (resulting in a boost of +0.7\% in C-Subject and up to +7.7\% in C-View) and part-aware LSTM \cite{Shahroudy2016} (achieving an improvement of +7\% in C-Subject and up to +15.1\% in C-View) while performing better than temporal CNN \cite{Kim2017} (up to +2.3\%) in C-View setting.
These results show that our unsupervised residual convolutions with Laplacian regularization exceed even supervised GRUs, RNNs, and LSTMs (and variants) for HAR.
Besides the mentioned favorable results of our AE-L, it is important to note that fully supervised techniques \cite{Zhang_2017_ICCV,Mengyuan2017,ChaoLi2017,yan2018spatial,Wen_Gao_Fu_Zhang_Xia_2019,LiChenChenZhang2019,shi2019,Si_2019_CVPR,Shi_2019_CVPR,cheng2020skeleton} perform better than our AE-L.
These methods mostly implement GCNs \cite{yan2018spatial,Wen_Gao_Fu_Zhang_Xia_2019,LiChenChenZhang2019,shi2019,Si_2019_CVPR,cheng2020skeleton}, and some of them additionally adapt LSTMs \cite{Si_2019_CVPR} or a variable temporal dense block \cite{Wen_Gao_Fu_Zhang_Xia_2019}.
The best performing method is \cite{cheng2020skeleton} with 90.7\% and 96.5\% in C-Subject and C-View, respectively.

%-------------------------------------------------------------------------
\vspace{-3 mm}
\subsection{Transfer across viewpoints for U-HAR} \label{section:transfer}
\begin{table}[t] \centering \resizebox{0.8\columnwidth}{!}{
    \begin{tabular}{llcccccc} \hline
        \multicolumn{2}{c}{\multirow{2}{*}{\bf U-HAR: Transfer Across Viewpoints}} & \multirow{2}{*}{\# of params.} & \multirow{2}{*}{where?} & NTU-60 & NTU-120 \\
        & & & & C-View & C-Setup \\ \hline
        Baseline & \cite{Su2020} & 0.58M & input (pre-proc) & 76.3\% & 42.7\% \\
        SeBiReNet & \cite{nie2020unsupervised} & 0.27M & input (data-aug) & 79.7\% & -- \\
        \textit{Our} {\bf GRAE} & (AE + SSVI) & \textit{0.39M} & \textit{feature space} & \textit{81.9\%} & \textit{47.0\%} \\
        \textit{Our} {\bf GRAE-L} & (AE + $\mathcal{R}_{\rm skel}$ + SSVI) & \textit{0.39M} & \textit{feature space} & \textbf{\textit{82.4\%}} & \textbf{\textit{48.9\%}} \\ \hline
    \end{tabular}}
    \vspace{2 mm}
    \caption{We compare our SSVI module, plugged into either AE and AE-L (Linear Evaluation Protocol \cite{zheng2018unsupervised}, our numbers in italic) with published results of \cite{Su2020,nie2020unsupervised}.}
    \vspace{-5mm}
    \label{table:GRL}
\end{table}

\vspace{-2 mm}
Since U-HAR is, by design, better tailored to real-world applications, we intended to push our approach to the limit and compete against SeBiReNet \cite{nie2020unsupervised} to transfer across viewpoints.
SeBiReNet and our GRAE-L (AE + $\mathcal{R}_{\rm skel}$ + SSVI) leverage random rotational noise to perturb the input data with a sharp algorithmic difference.
The two-stream Siamese architecture of SeBiReNet \cite{nie2020unsupervised} is jointly fed by rotated and non-rotated data while using non-adversarial optimization to promote viewpoints invariance.
Differently, we exploit gradient reversing \cite{ganin2015unsupervised} to achieve viewpoint invariance in a model which is fed by \emph{rotated data only}, attempting to fool a regressor (one ReLU-hidden layer MLP with a sigmoid readout layer) to predicting the triplet of Euler's angles used to rotate each mini-batch (see Algorithm \ref{algorithm:pseudo-code}).
We rely on a single stream, and as opposed to having two lightweight streams helping each other in generalizing better \cite{nie2020unsupervised}, our network is deeper (also depends upon a greater number of learnable parameters - 0.27M versus 0.39M, see Table \ref{table:GRL}) but achieves a better invariance across viewpoints.
Furthermore, our approach does not benefit from auxiliary skeletal datasets as commonly happening in unsupervised domain adaptation \cite{ganin2015unsupervised} (\eg, in SeBiReNet \cite{nie2020unsupervised}, a pre-training is performed on Cambridge-Imperial APE dataset, and then transfer learning is applied for NTU-60).
As seen in Table \ref{table:GRL}, our GRAE (AE+SSVI) and GRAE-L (AE- $\mathcal{R}_{\rm skel}$+SSVI) approaches score favorably against SeBiReNet \cite{nie2020unsupervised}, and GRAE-L has a +2.7\% on NTU-60 C-View setting.
In the same table, we also report a comparison with the baseline solution \cite{Su2020}, applying view-invariant transformations to "clean" the data from rotations as pre-processing.
Notably, despite our data being trained with more complex data to be fitted (our single-stream GRAE-L never sees non-rotated data), we still outperform this baseline by big margins (+6.1\% on NTU-60 \cite{Shahroudy_2016_NTURGBD} C-View and +6.2\% on NTU-120 \cite{Liu_2019_NTURGBD120} C-Setup).
\vspace{-3 mm}
%-------------------------------------------------------------------------
\subsection{Using synthetic data in training} \label{section:syntheticData}
\vspace{-2 mm}
This section investigates the impact of using synthetic data in training for C-View and C-Setup scenarios.
Synthetic data were obtained as described in Section \ref{section:GRL} (also Supplementary Material), and it was fixed for all experiments in Table \ref{table:synthetic}.
The models are trained with a) real data only, b) real + synthetic data, c) synthetic data only.
Real data refers to pre-processed data, so-called clean data in Section \ref{section:transfer}, which is already aligned to the same viewpoint.
\begin{table}[t] \centering \resizebox{0.9\columnwidth}{!}{
    \begin{tabular}[t]{llcccccccc} \hline
                                                        &                           & AE   & AE-L & GRAE (AE + SSVI) & GRAE-L (AE-L + SSVI) \\ \hline
                                                        & Real data (pre-processed)             & 85.1 & 85.4 & $\sim$           & $\sim$               \\
        NTU-60 \cite{Shahroudy_2016_NTURGBD} C-View & Real + Synthetic data & 80.4 & 80.6 & $\sim$           & $\sim$               \\
                                                        & Synthetic data        & 80.1 & 81.3 & \bf 81.9             & \bf 82.4                 \\ \hline
                                                        & Real data (pre-processed)            & 61.8 & 62.4 & $\sim$           & $\sim$               \\
        NTU-120 \cite{Liu_2019_NTURGBD120} C-Setup  &  Real + Synthetic data & 45.7 & 45.2 & $\sim$           & $\sim$               \\
                                                        &  Synthetic data        & 46.1 & 46.4 & \bf 47.0             & \bf 48.9                 \\ \hline
    \end{tabular}}
    \caption{{Performances (accuracy) of the proposed methods using the real and/or synthetic data in training. \textit{Notice that methods with SSVI rely only on synthetic data.}}}
    \vspace{-3 mm}
    \label{table:synthetic}
\end{table}
\begin{table}[t] \centering \resizebox{0.7\columnwidth}{!}{
    \begin{tabular}[t]{lcccc} \hline
        & \multicolumn{2}{c}{\bf NTU-60 \cite{Shahroudy_2016_NTURGBD}} & \multicolumn{2}{c}{\bf NTU-120 \cite{Liu_2019_NTURGBD120}} \\ 
                                        & \bf C-Subject & \bf C-View & \bf C-Subject & \bf C-Setup \\ \hline
        Our AE (Unsupervised)           & 52.3          & 81.0       & 41.0      & 44.5            \\
        Our AE End-to-end (Supervised)  & 69.8          & 83.7       & 57.1      & 59.6            \\
        Our AE Fine-tuning (Supervised) & 70.5          & 83.8       & 57.5      & 61.1            \\ \hline
    \end{tabular}}
    \caption{Performances of our AE (accuracy) in different learning schemes.}
    \vspace{-5mm}
    \label{table:finetune}
\end{table}
Recalling that SSVI-based experiments rely only on synthetic data, whose amount is as much as the real training data, the training set size of real + synthetic experiments is twice of real only and synthetic only.
Synthetic data includes rotational perturbations of \emph{not pre-processed} real data.
Thus, experiments only with synthetic data and real + synthetic data result in performance degradation for all models.
Experiments with real data perform the best out of all, but it is important to notice that \textit{the applied pre-processing is mostly not applicable in real-world applications as the viewpoints might not be known.}
When the amount of synthetic data in real + synthetic setting is decreased, performance increases, e.g., AE performs 83.2\% and 46.8\%, AE-L performs 84.3\% and 47.4\% on NTU60, and NTU120 with "real + (20\%)synthetic data".
In this case, SSVI-based models perform better than AE and AE-L (both synthetic \& real + synthetic) for all cases, showing that they can handle \emph{viewpoint perturbations} in a better way.
  
\vspace{-4mm}

%-------------------------------------------------------------------------
\subsection{Fine-tuning and end-to-end supervised training}
\label{fineTuning}
\vspace{-2 mm}
The performances of \textit{Our AE} with the fine-tuning protocol \cite{li2021crossclr} and end-to-end supervised training are reported in Table \ref{table:finetune}.
\textbf{Fine-tuning protocol} refers to first end-to-end pre-training of our AE in an unsupervised way and then appending a linear classifier to the encoder of AE, which is trained for HAR using the action labels.
It was applied for 100 epochs with learning rate 0.001.
\textbf{End-to-end training} refers to supervised training of our AE from scratch using the action labels of training data.
It was applied for 100 epochs with learning rate 0.001. While fine-tuning performs the best out of all, fine-tuning and supervised HAR results are always better than "Our AE Unsupervised" (as expected) by +3-18\% for NTU-60 \cite{Shahroudy_2016_NTURGBD} and +15-17\% for NTU-120 \cite{Liu_2019_NTURGBD120}.
As we train the Laplacian regularizer on the reconstructed skeleton from the \emph{decoder}, and the experiments presented herein are regarding applying a linear classifier appended to the \emph{encoder}, the results of \textit{Our AE-L} are the same as \textit{Our AE}.

%-------------------------------------------------------------------------
\vspace{-4.5mm}
\section{Conclusion} \label{section:conclusion}
\vspace{-3 mm}
We have introduced a novel unsupervised feature learning method that results in effective feature representations of actions from the input 3D skeleton sequences.
Our method is based on convolutional autoencoders (AE) and adapting Laplacian Regularization (L) to capturing the pose geometry in time.
Our AE-L is validated on large-scale HAR benchmarks where it exceeds all of SOTA skeleton-based U-HAR methods for cross-subject, cross-view, and cross-setup settings.
This proves that our AE-L is able to learn more distinctive action features compared to prior art.
We also upgrade AE-L with gradient reversing (GRAE-L) to provide better invariance to camera viewpoint changes compared to a direct competitor \cite{nie2020unsupervised}.
As future work, we will focus on enforcing the spatio-temporal connectivity through regularization over time and also concentrate on the real-time deployment of our AE-L.

%-------------------------------------------------------------------------
%\clearpage
\bibliography{refs}

\begin{thebibliography}{36}
\providecommand{\natexlab}[1]{#1}
\providecommand{\url}[1]{\texttt{#1}}
\expandafter\ifx\csname urlstyle\endcsname\relax
  \providecommand{\doi}[1]{doi: #1}\else
  \providecommand{\doi}{doi: \begingroup \urlstyle{rm}\Url}\fi

\bibitem[Cavazza et~al.(2019)Cavazza, Morerio, and Murino]{cavazza2019}
Jacopo Cavazza, Pietro Morerio, and Vittorio Murino.
\newblock Scalable and compact {3D} action recognition with approximated rbf
  kernel machines.
\newblock \emph{Pattern Recognition}, 93:\penalty0 25--35, 2019.

\bibitem[{Chao Li} et~al.(2017){Chao Li}, {Qiaoyong Zhong}, {Di Xie}, and
  {Shiliang Pu}]{ChaoLi2017}
{Chao Li}, {Qiaoyong Zhong}, {Di Xie}, and {Shiliang Pu}.
\newblock Skeleton-based action recognition with convolutional neural networks.
\newblock In \emph{2017 IEEE International Conference on Multimedia Expo
  Workshops (ICMEW)}, pages 597--600, 2017.
\newblock \doi{10.1109/ICMEW.2017.8026285}.

\bibitem[Cheng et~al.(2020{\natexlab{a}})Cheng, Zhang, He, Chen, Cheng, and
  Lu]{Cheng_2020_CVPR}
Ke~Cheng, Yifan Zhang, Xiangyu He, Weihan Chen, Jian Cheng, and Hanqing Lu.
\newblock Skeleton-based action recognition with shift graph convolutional
  network.
\newblock In \emph{Proceedings of the IEEE/CVF Conference on Computer Vision
  and Pattern Recognition (CVPR)}, June 2020{\natexlab{a}}.

\bibitem[Cheng et~al.(2020{\natexlab{b}})Cheng, Zhang, He, Chen, Cheng, and
  Lu]{cheng2020skeleton}
Ke~Cheng, Yifan Zhang, Xiangyu He, Weihan Chen, Jian Cheng, and Hanqing Lu.
\newblock Skeleton-based action recognition with shift graph convolutional
  network.
\newblock In \emph{Proceedings of the IEEE/CVF Conference on Computer Vision
  and Pattern Recognition}, pages 183--192, 2020{\natexlab{b}}.

\bibitem[Deo(1974)]{deo1974graph}
Narsingh Deo.
\newblock \emph{Graph theory with applications to engineering and computer
  science}.
\newblock Courier Dover Publications, 1974.

\bibitem[Ganin and Lempitsky(2015)]{ganin2015unsupervised}
Yaroslav Ganin and Victor Lempitsky.
\newblock Unsupervised domain adaptation by backpropagation.
\newblock In \emph{The International Conference on Machine Learning (ICML)},
  2015.

\bibitem[Holden et~al.(2015)Holden, Saito, Komura, and
  Joyce]{holden2015Siggraph}
Daniel Holden, Jun Saito, Taku Komura, and Thomas Joyce.
\newblock Learning motion manifolds with convolutional autoencoders.
\newblock In \emph{SIGGRAPH Asia 2015 Technical Briefs}, 2015.

\bibitem[{Kim} and {Reiter}(2017)]{Kim2017}
T.~S. {Kim} and A.~{Reiter}.
\newblock Interpretable 3d human action analysis with temporal convolutional
  networks.
\newblock In \emph{2017 IEEE Conference on Computer Vision and Pattern
  Recognition Workshops (CVPRW)}, pages 1623--1631, 2017.
\newblock \doi{10.1109/CVPRW.2017.207}.

\bibitem[Kundu et~al.(2018)Kundu, Gor, Uppala, and Babu]{kundu2018unsupervised}
Jogendra~Nath Kundu, Maharshi Gor, Phani~Krishna Uppala, and R~Venkatesh Babu.
\newblock Unsupervised feature learning of human actions as trajectories in
  pose embedding manifold.
\newblock In \emph{IEEE Winter Conference on Applications of Computer Vision
  (WACV).}, 2018.

\bibitem[Li et~al.(2018)Li, Wong, Zhao, and Kankanhalli]{li2018unsupervised}
Junnan Li, Yongkang Wong, Qi~Zhao, and Mohan~S Kankanhalli.
\newblock Unsupervised learning of view-invariant action representations.
\newblock \emph{Advances in Neural Information Processing Systems (NeurIPS)},
  2018.

\bibitem[{Li} et~al.(2019){Li}, {Chen}, {Chen}, {Zhang}, {Wang}, and
  {Tian}]{LiChenChenZhang2019}
M.~{Li}, S.~{Chen}, X.~{Chen}, Y.~{Zhang}, Y.~{Wang}, and Q.~{Tian}.
\newblock Actional-structural graph convolutional networks for skeleton-based
  action recognition.
\newblock In \emph{2019 IEEE/CVF Conference on Computer Vision and Pattern
  Recognition (CVPR)}, pages 3590--3598, 2019.
\newblock \doi{10.1109/CVPR.2019.00371}.

\bibitem[Lin et~al.(2020)Lin, Song, Yang, and Liu]{lin2020ms2l}
Lilang Lin, Sijie Song, Wenhan Yang, and Jiaying Liu.
\newblock Ms2l: Multi-task self-supervised learning for skeleton based action
  recognition.
\newblock In \emph{Proceedings of the 28th ACM International Conference on
  Multimedia}, pages 2490--2498, 2020.

\bibitem[Linguo et~al.(2021)Linguo, Minsi, Bingbing, Hang, Jiancheng, and
  Wenjun]{li2021crossclr}
Li~Linguo, Wang Minsi, Ni~Bingbing, Wang Hang, Yang Jiancheng, and Zhang
  Wenjun.
\newblock 3d human action representation learning via cross-view consistency
  pursuit.
\newblock In \emph{CVPR}, 2021.

\bibitem[Liu et~al.(2016)Liu, Shahroudy, Xu, and Wang]{liu2016spatiotemporal}
Jun Liu, Amir Shahroudy, Dong Xu, and Gang Wang.
\newblock Spatio-temporal lstm with trust gates for 3d human action
  recognition.
\newblock \emph{arXiv preprint arXiv:1607.07043}, 2016.

\bibitem[Liu et~al.(2019)Liu, Shahroudy, Perez, Wang, Duan, and
  Kot]{Liu_2019_NTURGBD120}
Jun Liu, Amir Shahroudy, Mauricio Perez, Gang Wang, Ling-Yu Duan, and Alex~C.
  Kot.
\newblock Ntu rgb+d 120: A large-scale benchmark for 3d human activity
  understanding.
\newblock \emph{IEEE Transactions on Pattern Analysis and Machine
  Intelligence}, 2019.
\newblock \doi{10.1109/TPAMI.2019.2916873}.

\bibitem[Liu et~al.(2017)Liu, Liu, and Chen]{Mengyuan2017}
Mengyuan Liu, Hong Liu, and Chen Chen.
\newblock Enhanced skeleton visualization for view invariant human action
  recognition.
\newblock \emph{Pattern Recogn.}, 68:\penalty0 346–362, August 2017.
\newblock \doi{10.1016/j.patcog.2017.02.030}.

\bibitem[Mikhail~Belkin(2006)]{belkin2006manifoldreg}
Vikas~Sindhwani Mikhail~Belkin, Partha~Niyogi.
\newblock Manifold regularization: A geometric framework for learning from
  labeled and unlabeled examples.
\newblock \emph{International Journal of Machine Learning Research (JMLR)},
  7\penalty0 (11), 2006.

\bibitem[Nie et~al.(2020)Nie, Liu, and Liu]{nie2020unsupervised}
Qiang Nie, Ziwei Liu, and Yunhui Liu.
\newblock Unsupervised human 3d pose representation with viewpoint and pose
  disentanglement.
\newblock In \emph{Springer European Conference on Computer Vision (ECCV)},
  2020.

\bibitem[{Pang} and {Cheung}(2017)]{pang2017denoisingAE}
J.~{Pang} and G.~{Cheung}.
\newblock Graph laplacian regularization for image denoising: Analysis in the
  continuous domain.
\newblock \emph{IEEE Transactions on Image Processing}, 26\penalty0
  (4):\penalty0 1770--1785, 2017.

\bibitem[Paoletti et~al.(2020)Paoletti, Cavazza, Beyan, and
  Bue]{paoletti2020subspace}
Giancarlo Paoletti, Jacopo Cavazza, Cigdem Beyan, and Alessio~Del Bue.
\newblock Subspace clustering for action recognition with covariance
  representations and temporal pruning.
\newblock In \emph{Proceedings of the International Conference on Pattern
  Recognition (ICPR)}, 2020.

\bibitem[{Rahmani} et~al.(2016){Rahmani}, {Mahmood}, {Huynh}, and
  {Mian}]{Rahmani2016}
H.~{Rahmani}, A.~{Mahmood}, D.~{Huynh}, and A.~{Mian}.
\newblock Histogram of oriented principal components for cross-view action
  recognition.
\newblock \emph{IEEE Transactions on Pattern Analysis and Machine
  Intelligence}, 38\penalty0 (12):\penalty0 2430--2443, 2016.
\newblock \doi{10.1109/TPAMI.2016.2533389}.

\bibitem[Rao et~al.(2020)Rao, Xu, Hu, Cheng, and Hu]{rao2020augmented}
Haocong Rao, Shihao Xu, Xiping Hu, Jun Cheng, and Bin Hu.
\newblock Augmented skeleton based contrastive action learning with momentum
  lstm for unsupervised action recognition, 2020.

\bibitem[{Shahroudy} et~al.(2016){Shahroudy}, {Liu}, {Ng}, and
  {Wang}]{Shahroudy2016}
A.~{Shahroudy}, J.~{Liu}, T.~{Ng}, and G.~{Wang}.
\newblock Ntu rgb+d: A large scale dataset for 3d human activity analysis.
\newblock In \emph{2016 IEEE Conference on Computer Vision and Pattern
  Recognition (CVPR)}, pages 1010--1019, 2016.
\newblock \doi{10.1109/CVPR.2016.115}.

\bibitem[Shahroudy et~al.(2016)Shahroudy, Liu, Ng, and
  Wang]{Shahroudy_2016_NTURGBD}
Amir Shahroudy, Jun Liu, Tian-Tsong Ng, and Gang Wang.
\newblock Ntu rgb+d: A large scale dataset for 3d human activity analysis.
\newblock In \emph{IEEE Conference on Computer Vision and Pattern Recognition},
  June 2016.

\bibitem[Shi et~al.(2019{\natexlab{a}})Shi, Zhang, Cheng, and
  Lu]{Shi_2019_CVPR}
Lei Shi, Yifan Zhang, Jian Cheng, and Hanqing Lu.
\newblock Skeleton-based action recognition with directed graph neural
  networks.
\newblock In \emph{Proceedings of the IEEE/CVF Conference on Computer Vision
  and Pattern Recognition (CVPR)}, June 2019{\natexlab{a}}.

\bibitem[Shi et~al.(2019{\natexlab{b}})Shi, Zhang, Cheng, and Lu]{shi2019}
Lei Shi, Yifan Zhang, Jian Cheng, and Hanqing Lu.
\newblock Two-stream adaptive graph convolutional networks for skeleton-based
  action recognition.
\newblock In \emph{CVPR}, 2019{\natexlab{b}}.

\bibitem[Si et~al.(2019)Si, Chen, Wang, Wang, and Tan]{Si_2019_CVPR}
Chenyang Si, Wentao Chen, Wei Wang, Liang Wang, and Tieniu Tan.
\newblock An attention enhanced graph convolutional lstm network for
  skeleton-based action recognition.
\newblock In \emph{Proceedings of the IEEE/CVF Conference on Computer Vision
  and Pattern Recognition (CVPR)}, June 2019.

\bibitem[{Su} et~al.(2020){Su}, {Liu}, and {Shlizerman}]{Su2020}
K.~{Su}, X.~{Liu}, and E.~{Shlizerman}.
\newblock Predict \& cluster: Unsupervised skeleton based action recognition.
\newblock In \emph{2020 IEEE/CVF Conference on Computer Vision and Pattern
  Recognition (CVPR)}, pages 9628--9637, 2020.
\newblock \doi{10.1109/CVPR42600.2020.00965}.

\bibitem[Wen et~al.(2019)Wen, Gao, Fu, Zhang, and
  Xia]{Wen_Gao_Fu_Zhang_Xia_2019}
Yu-Hui Wen, Lin Gao, Hongbo Fu, Fang-Lue Zhang, and Shihong Xia.
\newblock Graph cnns with motif and variable temporal block for skeleton-based
  action recognition.
\newblock \emph{Proceedings of the AAAI Conference on Artificial Intelligence},
  33\penalty0 (01):\penalty0 8989--8996, Jul. 2019.
\newblock \doi{10.1609/aaai.v33i01.33018989}.
\newblock URL \url{https://ojs.aaai.org/index.php/AAAI/article/view/4929}.

\bibitem[Xu et~al.(2020)Xu, Rao, Hu, and Hu]{xu2020prototypical}
Shihao Xu, Haocong Rao, Xiping Hu, and Bin Hu.
\newblock Prototypical contrast and reverse prediction: Unsupervised skeleton
  based action recognition.
\newblock In \emph{arXiv preprint 2011.07236}, 2020.

\bibitem[Yan et~al.(2018)Yan, Xiong, and Lin]{yan2018spatial}
Sijie Yan, Yuanjun Xiong, and Dahua Lin.
\newblock Spatial temporal graph convolutional networks for skeleton-based
  action recognition.
\newblock \emph{AAAI}, pages 7444--7452, 2018.

\bibitem[Yang et~al.(2020)Yang, Li, Fu, Fan, and Leung]{yang2020centrality}
Dong Yang, Monica~Mengqi Li, Hong Fu, Jicong Fan, and Howard Leung.
\newblock Centrality graph convolutional networks for skeleton-based action
  recognition.
\newblock In \emph{European Conference on Computer Vision (ECCV)}, 2020.

\bibitem[{Yong Du} et~al.(2015){Yong Du}, {Wang}, and {Wang}]{Du2015}
{Yong Du}, W.~{Wang}, and L.~{Wang}.
\newblock Hierarchical recurrent neural network for skeleton based action
  recognition.
\newblock In \emph{2015 IEEE Conference on Computer Vision and Pattern
  Recognition (CVPR)}, pages 1110--1118, 2015.
\newblock \doi{10.1109/CVPR.2015.7298714}.

\bibitem[Zhang et~al.(2017)Zhang, Lan, Xing, Zeng, Xue, and
  Zheng]{Zhang_2017_ICCV}
Pengfei Zhang, Cuiling Lan, Junliang Xing, Wenjun Zeng, Jianru Xue, and Nanning
  Zheng.
\newblock View adaptive recurrent neural networks for high performance human
  action recognition from skeleton data.
\newblock In \emph{Proceedings of the IEEE International Conference on Computer
  Vision (ICCV)}, Oct 2017.

\bibitem[Zheng et~al.(2018)Zheng, Wen, Liu, Long, Dai, and
  Gong]{zheng2018unsupervised}
Nenggan Zheng, Jun Wen, Risheng Liu, Liangqu Long, Jianhua Dai, and Zhefeng
  Gong.
\newblock Unsupervised representation learning with long-term dynamics for
  skeleton based action recognition.
\newblock In \emph{Proceedings of the AAAI Conference on Artificial
  Intelligence (AAAI)}, 2018.

\bibitem[Zunino et~al.(2020)Zunino, Cavazza, Volpi, Morerio, Cavallo, Becchio,
  and Murino]{zunino2020predicting}
Andrea Zunino, Jacopo Cavazza, Riccardo Volpi, Pietro Morerio, Andrea Cavallo,
  Cristina Becchio, and Vittorio Murino.
\newblock Predicting intentions from motion: The subject-adversarial adaptation
  approach.
\newblock \emph{International Journal of Computer Vision}, 128\penalty0
  (1):\penalty0 220--239, 2020.

\end{thebibliography}

%-------------------------------------------------------------------------
\end{document}